\documentclass[sigconf]{acmart}

\usepackage{amsmath}
\usepackage{algorithm}
\usepackage{algpseudocode}
\usepackage{booktabs}
\AtBeginDocument{%
  }

\acmBooktitle{The 10th Mining and Learning from Time Series Workshop: From Classical Methods to LLMs}

\begin{document}

\title{Time series forecasting with high stakes: A field study of the air cargo industry}


\author{Abhinav Garg}
\affiliation{%
  \institution{FLYR Inc.}
  \city{San Francisco}
  \country{CA}}
\email{abhinav.garg@flyr.com}

\author{Naman Shukla}
\affiliation{%
  \institution{FLYR Inc.}
  \city{Gainesville}
  \country{FL}}
\email{naman.shukla@flyr.com}

\author{Maarten Wormer}
\affiliation{%
  \institution{FLYR Inc.}
  \city{Amsterdam}
  \country{NL}}
\email{maarten.wormer@flyr.com}

\renewcommand{\shortauthors}{Garg et al.}

\begin{abstract}
  Time series forecasting in the air cargo industry presents unique challenges due to volatile market dynamics and the significant impact of accurate forecasts on generated revenue. This paper explores a comprehensive approach to demand forecasting at the origin-destination (O\&D) level, focusing on the development and implementation of machine learning models in decision-making for the air cargo industry. We leverage a mixture of experts framework, combining statistical and advanced deep learning models to provide reliable forecasts for cargo demand over a six-month horizon. The results demonstrate that our approach outperforms industry benchmarks, offering actionable insights for cargo capacity allocation and strategic decision-making in the air cargo industry. While this work is applied in the airline industry, the methodology is broadly applicable to any field where forecast-based decision-making in a volatile environment is crucial. 
\end{abstract}



\keywords{Time Series Forecasting, Mixture of Experts Framework, Revenue Management, Seasonal Trends, Deep Learning, Meta-Learning, Case Study}


\maketitle

\section{Introduction}
The increasing availability and complexity of time series data have amplified the need for robust forecasting techniques across various industries, including finance, healthcare, and transportation \cite{lim2021time, sezer2020financial, torres2021deep}. In the air cargo sector, accurate forecasting is critical for maximizing revenue generated through each unit of capacity used by cargo bookings \cite{kasilingam1997air, billings2003cargo}. Unlike passenger sales, air cargo bookings are notoriously volatile, are prone to be cancelled, and are influenced by many factors such as economic conditions, seasonal events, and disruptions affecting global trade \cite{kupfer2017underlying}. Additionally, cargo bookings are often part of longer-term customer commitments, which help airlines plan their network schedules and budgeting efforts. Finally, since there are significantly fewer cargo bookings per flight as compared to passenger bookings, the importance of making the right decision (informed by an accurate forecast) is increased. This paper presents a field study on time series forecasting in the air cargo industry, focusing on the challenges and solutions associated with mid-to-long-term demand prediction.

Recent advancements in sensing technologies and data analytics have paved the way for more sophisticated forecasting models \cite{heng2009study, anguita2023air}. However, the air cargo industry faces unique challenges, including irregular booking patterns, varying shipment types, and the need for high accuracy in forecasts. Traditional forecasting methods often fail to address these complexities, necessitating the development of specialized machine-learning approaches.

In this study, we introduce a comprehensive forecasting framework that combines multiple statistical and machine learning models to predict cargo demand at the origin and destination level. Our approach involves a detailed data preparation process, integrating various data sources to capture the full spectrum of factors affecting cargo demand. We employ a mixture of experts framework, which allows us to select the best-performing model for each sample based on historical performance, thereby enhancing the overall prediction accuracy.

Our research is conducted within the context of a major air cargo carrier, providing a real-world application of the proposed method, and validated across diverse market segments. We evaluate the performance of our models using key metrics such as root mean square error (RMSE) and weighted normalised root mean square (WnRMSE), demonstrating significant improvements over existing industry benchmark forecasting techniques. The findings of this study offer valuable insights for applied teams in various industries, enabling more informed decision-making leading towards better adoption of forecasting models.

Our contributions are as follows:
\begin{itemize}
    \item We present an applied real-world case study where time-series forecasting is crucial in decision-making. Additionally, we describe the modeling process to make machine learning solution adoption smoother by the end user. 
    \item We propose a practically usable framework of a mixture of experts. Also, our training methodology of meta-learning on machine learning models outperformed industry benchmarks. 
    \item We share results from our implementation backtested on real data from a major airline carrier, aiming to make high-stakes decisions that significantly impact the majority of the annual revenue.
\end{itemize}

\subsection{Background}
In the context of airline cargo, mid-term refers to a forecasting period of up to six months ahead, effectively covering everything beyond the typical cargo booking window of up to 30 days before the flight departure. Mid-term forecasting is crucial because it allows stakeholders to have a proper idea of upcoming trends and demands, enabling them to make informed decisions on mid-term allotment requests. Mid-term allotment contracts in air cargo are agreements between airlines and shippers or freight forwarders to allocate a certain amount of cargo space on flights over a defined period, typically ranging from a few months to a couple of years \cite{sales2023air, hellermann2006capacity, kasilingam1997air}. These contracts facilitate more accurate demand forecasting and reduce risks associated with fluctuating cargo volumes and market rates \cite{pun2010air, amaruchkul2011note}. For shippers and freight forwarders, these contracts provide guaranteed access to cargo space, enhancing service reliability and customer satisfaction, especially during peak seasons or unexpected demand surges \cite{popescu2010air, zivcic2022applicability}. The contracts typically detail the volume of cargo space, routes, duration, and pricing terms, with negotiations often considering historical cargo volumes, projected future needs, and current market conditions \cite{amaruchkul2011note}. Despite the advantages, challenges arise from the dynamic nature of global trade, economic fluctuations, and unexpected events like the COVID-19 pandemic, which impact cargo volumes and market rates. Flexibility and adaptive strategies are essential in managing these contracts amidst such uncertainties \cite{becker2007managing, dewulf2014strategy}. Overall, mid-term allotment contracts are pivotal in ensuring stability and predictability in the air cargo industry, benefiting both airlines and their customers by providing a structured approach to capacity allocation and revenue management \cite{feng2015tying, wada2017risk}.

\subsection{Related Work}

Forecasting techniques have significantly evolved, with traditional methods like ARIMA and exponential smoothing remaining popular due to their simplicity and effectiveness in stable environments \cite{de200625}. Recent advancements in computational capabilities have led to the development of more sophisticated models, particularly neural network-based approaches. These models, including convolutional neural networks (CNNs) and recurrent neural networks (RNNs), have demonstrated superior performance in capturing complex, non-linear relationships in data, thereby enhancing forecasting accuracy \cite{kim2019forecasting, siami2018comparison}. Additionally, attention-based neural network architectures have also shown some promise in time series forecasting \cite{shih2019temporal, lim2021temporal}. 

Applications of such techniques in airline revenue management have picked up recently. For instance, application of machine learning-based solutions for dynamic pricing and passenger flow estimation \cite{koc2021dynamic, yu2022short} in passenger business. Recent research on gravity models, dynamic linear model, fuzzy regression forecasting models and simplified variants of neural networks have shown promising results in air cargo industry \cite{alexander2021applications, rodriguez2020air, chou2011application, chen2012improving}. Historically, there have been fewer publications on air cargo demand forecasting, despite its considerable importance, especially when compared to the extensive research on air passenger demand forecasting \cite{anguita2023air}. 



\section{Decision factors}

In this section, we explain the two primary factors that are responsible for mid-term allotment decision-making: demand estimation and, airline network and capacity. 

\subsection{Demand estimation}
Estimating the upcoming mid-term demand is crucial for airline cargo carriers as it allows them to allocate capacity efficiently between long-term stable contracts and short-term volatile bookings, which typically arrive within 30 days. Demand for air cargo is influenced by various factors including the origin and destination, product category, global events (i.e., COVID-19 and Suez Canal blockade), and macro trends \cite{kupfer2017underlying}. For instance, countries like Columbia and Ethiopia are significant exporters of flowers, resulting in a peak in cargo volumes before Valentine's Day and Mother's Day. Similarly, there is a notable increase in import volumes of electronics and technology components from Asia to Europe and North America during Black Friday and Cyber Monday. Moreover, macroeconomic policies and trends play a critical role in influencing both the demand and supply of products.

\subsection{Network and capacity}
Passenger airlines with significant revenue from air cargo typcially operate a hub-and-spoke network model \cite{brueckner1992fare}, where flights are routed through a central hub to various destinations. This model facilitates efficient passenger transfers and operational management. In contrast, air cargo logistics often prioritize the origin and destination points without significant regard to intermediary stops, focusing instead on cost-effectiveness and efficient routes for their shipments. To optimize the overall logistics, a comprehensive network-wide consideration of capacity is essential. This includes evaluating forward-looking schedules and sellable capacities, encompassing both passenger and cargo flights for each flight leg \cite{levina2011network}. Such a holistic approach ensures that both passenger and cargo needs are met efficiently, maximizing the use of available resources and improving overall service delivery.

\section{Forecasting framework}

We want to predict the total demand in terms of expected shipped weight, volume, and revenue for each week for each origin and destination (O\&D) for the next 6 months. Airline carriers may have a lot of cargo destinations that they fly to and therefore, the possible O\&D combinations for forecasting could be in the order of tens of thousands. To make the problem simpler, we only focus on O\&D that contribute to the top 90\% of cargo revenue for the airline. We call this group of O\&Ds the significant cluster. This reduces the problem space to a few 1000s of O\&Ds. Through exploratory data analysis, we found that most of the variability and important patterns in the data are captured by the high-frequency signals at the departure date level. Therefore in our approach, we make predictions at the O\&D - departure date level, and then aggregate them up to the O\&D - departure week level.

\subsection{Models}

We experimented with eight different statistical models for baseline analysis. Additionally, the industry-wide benchmark, used by the majority of airlines, relies on year-over-year predictions. We used Croston Classic, Historic Average, Auto ETS, Window Average, Dynamic Optimized Theta (DOT), Simple Exponential Smoothing (SES), Holt-Winters, and Seasonal Naive for statistical baseline \cite{croston1972forecasting, jain2017study, fioruci2015optimised, chatfield1978holt, hyndman2018forecasting}. With their strengths and weaknesses, these methods collectively aim to capture a broad range of trend and seasonality patterns in the time series that help set a baseline for building advanced models. Furthermore, our study focused on the implementation of three machine learning model architectures.

\subsubsection{DNN-LADD}
\begin{figure}[h]
    \centering
    \includegraphics[width=\linewidth]{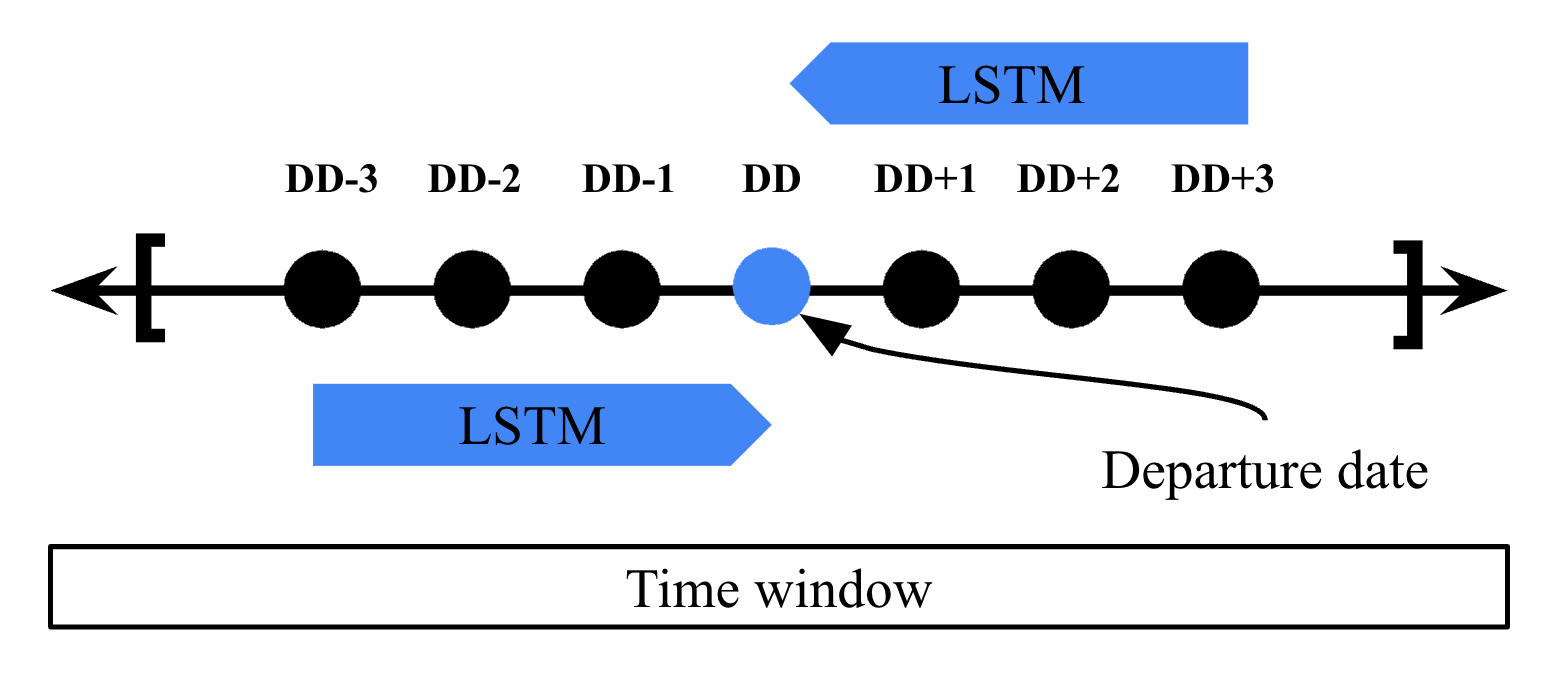}
    \caption{Illustration of look around departure date (LADD) method using the bidirectional LSTM }
    \label{fig:ladd}
\end{figure}

We implemented a simplified deep neural network (DNN) based primarily on fully connected layers assuming independent and identically distributed data points as inputs. Each timestamp point is considered independent of others without considering the sequential nature of the time series. The only time series considered is a small period that "Looks Around the Departure Dates" (LADD). These features help capture the static information in the time series and any seasonal and calendar events that the airline cares about. The LADD features are currently set to be equal to harmonic features \cite{panofsky1958some} and are encoded using a bidirectional LSTM \cite{hochreiter1997long} as illustrated in the figure \ref{fig:ladd}. All categorical features go through their embedding layers. These embeddings and encodings along with the dense features are concatenated together before being fed to the fully connected layers. The primary hypothesis for implementing this architecture is to capture patterns based on context rather than driven by traditional time series concepts, leveraging the flexibility of neural networks to model complex relationships within the data.

\subsubsection{NBEATS}
This is a state-of-the-art deep-learning-based time-series forecasting model proven to work well on short-horizon forecasting \cite{oreshkin2019n}. It takes a time-series input to predict the entire forecasting horizon in one shot. The model tries to sequentially learn the residual from the inputs with the use of stacks. By default, NBEATS is a univariate time-series model, that is, it only takes time-series being forecasted as input. To incorporate additional input features, we modified its architecture to add context in the form of encodings concatenated together to the input vector. We observed that this model performs better when patterns are seasonal and struggles when data is sparse or highly context-driven such as capacity or events. LADD and embedding features are learned the same way as the DNN-LADD model. Dense features are embedded using an attention block \cite{vaswani2017attention}. 

\subsubsection{TFT}
Temporal Fusion Transformer (TFT) is one of the transformational methods in time-series forecasting utilizing the transformer architecture \cite{lim2021temporal}. Unlike NBEATS, TFT is designed to support input features and can learn from static metadata, time-varying past inputs, and time-varying a priori known future inputs. It incorporates LSTM layers and a multi-head self-attention transformer block to enable efficient information flow, enhanced by skip connections. However, being a transformer-based architecture, this model is computationally expensive and requires significant memory to train, especially for long-horizon forecasting. 

\subsection{Meta learning}

Meta-learning is an advanced approach in artificial intelligence aimed at training systems to effectively learn new tasks or adapt to novel situations based on past experiences, rather than merely mastering specific tasks. This paradigm, often described as "learning to learn," enables a system to acquire new concepts or tasks with minimal additional training quickly. Given the variability in data sizes across different Origin-Destination (O\&D) pairs, traditional machine learning models can easily overfit to O\&Ds with larger datasets. Meta-learning addresses this issue by implementing a 2-step optimization process rather than standard gradient descent, as detailed in the Model-Agnostic Meta-Learning (MAML) framework \cite{finn2017maml}.

In this approach, each O\&D is treated as a distinct task, and gradients for model updates are derived by moving toward the local minima for a set of O\&Ds. This is achieved through a stochastic gradient descent update for a particular O\&D on a model copy in the inner loop using the support set. Concurrently, the gradients for the main model updates are gathered using a query set from the updated model copy in the outer loop. During inference, the model undergoes fine-tuning using a subset of each O\&D’s training data for a few steps before making predictions. A significant advantage of this meta-learning approach is its applicability to O\&Ds with limited historical data, such as the bottom 10\% of O\&Ds. By fine-tuning the model with minimal steps, accurate predictions can be made without extensive retraining, provided the data supports the creation of such O\&Ds.

\begin{algorithm}[]
\caption{O\&D Meta-Learning}
\label{alg:maml}
\begin{algorithmic}[1]
\Require $p(\mathcal{D})$: distribution over O\&D
\Require $\alpha, \beta$: step size hyperparameters
\State Randomly initialize $\theta$
\While{not done}
    \State Sample batch of O\&D: $\mathcal{D}_i \sim p(\mathcal{D})$
    \ForAll{$\mathcal{D}_i$}
        \State Evaluate $\nabla_\theta \mathcal{L}_{\mathcal{D}_i}(f_\theta)$ using support set
        \State Compute adapted parameters with gradient descent: $\theta'_i = \theta - \alpha \nabla_\theta \mathcal{L}_{\mathcal{D}_i}(f_\theta)$
    \EndFor
    \State Update $\theta \leftarrow \theta - \beta \nabla_\theta \sum_{\mathcal{D}_i \sim p(\mathcal{D})} \mathcal{L}_{\mathcal{D}_i}(f_{\theta'_i})$ using query set
\EndWhile
\end{algorithmic}
\end{algorithm}

In the context of the meta-learning algorithm, let \( p(\mathcal{D}) \) represent the distribution over tasks, where each O\&D (Origin-Destination pair) is considered a separate task. The step size hyperparameters, \( \alpha \) and \( \beta \), are employed in the inner and outer loop updates respectively, governing the magnitude of parameter adjustments during gradient descent. The model parameters are denoted by \( \theta \), while \( \theta'_i \) refers to the adapted parameters for a specific task \( \mathcal{D}_i \) after the inner loop update. The loss function for a task \( \mathcal{D}_i \) is represented by \( \mathcal{L}_{\mathcal{D}_i} \), which evaluates the model's performance on that task. Finally, \( f_\theta \) signifies the model parameterized by \( \theta \), encapsulating the function learned by the meta-learning algorithm. These variables collectively enable the systematic optimization and fine-tuning of the model across diverse tasks as mentioned in algorithm \ref{alg:maml}.

\subsection{Mixture of experts}
For each O\&D pair \(\mathcal{D}_i\), the model \(f_\theta\) with the lowest loss value \(\mathcal{L}\) on the validation set is selected, as denoted in equation \ref{eq:moe1}. The selected model \(f_{\theta_{\mathcal{D}}^*}\) is then used to make predictions on the test set for the corresponding O\&D pair. By selecting the best-performing model for each O\&D pair based on the validation loss, the overall prediction error across all O\&D pairs is minimized. This approach leverages the strengths of different models, acknowledging that no single model consistently performs best across all O\&D pairs. The implementation of this framework involves training each model on the training set, evaluating their performance on the validation set to compute the loss for each O\&D pair, and then selecting and using the model with the lowest loss for making predictions on the test set. This process ensures that the forecasting approach is tailored to the specific characteristics of each O\&D pair, ultimately aiming to minimize the overall prediction error through a data-driven model selection process.

\begin{equation}
\label{eq:moe1}
f_{\theta_{\mathcal{D}_i}^*} = \arg\min_{f_\theta \in \mathcal{F}} \mathcal{L}_{\mathcal{D}_i}(f_\theta)
\end{equation}

We chose the mixture of experts approach to provide the most accurate and reliable forecasts for air cargo analysts, thereby enabling them to make well-informed decisions. This method allows us to leverage baseline forecasts for inference, particularly in scenarios where the existing modeling approaches fall short, ensuring comprehensive coverage across various origin and destination pairs. Furthermore, this approach is apt for applications with low-frequency inference due to high computational complexity and hence affordable to mid-term forecasts as it needs to be updated only a few times a month. Therefore, the mixture of experts framework is a practically viable solution, aligning with the operational needs.

\section{Results}

\begin{table}[t]
\caption{Model performance and win ratio on significant cluster}
\label{table1}
\begin{tabular}{@{}lllll@{}}
\toprule
Model Name   & \begin{tabular}[c]{@{}l@{}}Model\\ Type\end{tabular}   & Win ratio & nRMSE & \begin{tabular}[c]{@{}l@{}}WnRMSE\end{tabular} \\ \midrule
TFT            & ML    & 0.245     &    1.493   &   1.042                                                       \\ 
Auto ETS       & Stat     & 0.160     &  1.459     &     1.016                                                     \\
NBEATS         & ML       & 0.151     &    1.440   &    1.000                                                      \\
DOT & Stat & 0.087     &   1.447    &    1.006                                                      \\
DNN-LADD       & ML       & 0.084     &   1.618    &     1.092                                                    \\
Holt-Winters   & Stat      & 0.070     &   1.698    &      1.178                                                    \\
SES & Stat  & 0.068     &  1.001     &     1.037                                                     \\
Window Average   & Stat     & 0.051     &   1.521    &      1.044                                                  \\
Historic Average  & Stat   & 0.044     &   1.626    &      1.094                                                    \\
Croston Classic & Stat    & 0.027     &  1.489     &     1.030                                                     \\
Seasonal Naive  & Stat    & 0.006     &   1.898    &      1.210                                                  \\

\bottomrule
\end{tabular}
\end{table}

The target variable chosen for all experiments is the total shipped cargo weight in kilograms on a given date for a given O\&D pair. All models are trained from April 2019 to March 2023, and validated on a period starting from April 2023 to September 2023 \footnote{The exact dates and revenue figures cannot be included due to proprietary, privacy, and sensitivity restrictions}.

The primary metrics chosen for evaluation are root mean squared error (RMSE), unweighted normalized root mean squared error (nRMSE), and weighted normalized root mean squared error (WnRMSE). The WnRMSE formulation is described in equation \ref{eq:nrmse}. 

\begin{equation}
\label{eq:nrmse}
    \textrm{WnRMSE} = \sum_{OD} w_{OD} \times \text{nRMSE}_{OD} 
\end{equation}
where $w_{OD} = \frac{sum(\text{weight})_{OD}}{\sum_{OD} sum(\text{weight})_{OD}},  
    \textrm{  nRMSE}_{OD} = \frac{\text{RMSE}_{OD}}{avg(\text{weight})_{OD}} $
\newline
    
RMSE, by definition, is biased towards high-value data points, leading to higher magnitude values for bigger O\&Ds in the airline network and potentially skewing network-wide RMSE values. Normalized RMSE (nRMSE) addresses this bias by normalizing RMSE values by the average of the target variable. Therefore, we calculate nRMSE for each O\&D to ensure fair performance comparison across the network. We use a weighted mean of the individual nRMSE values to obtain the network-wide error, where weights are the total shipped weight of each O\&D pair over the evaluation period. This approach provides a performance assessment of the overall network with proportionate relevance of O\&Ds.

Individual model performance is presented in Table \ref{table1}. Here, average nRMSE and WnRMSE are recorded for each model across the entire validation period. The win ratios are calculated for the significant cluster due to resource constraints. Advanced techniques like TFT and NBEATS are 2 of the top 3 best performing models within the significant cluster. Also, statistical methods like Auto ETS effectively capture seasonal trends, resulting in improved performance. 

\begin{table}[H]
\caption{Aggregated metrics for clustered O\&Ds}
\label{table2}
\begin{tabular}{@{}llllll@{}}
\toprule
O\&D's  & \begin{tabular}[c]{@{}l@{}}Sample \\ Size \end{tabular} & \begin{tabular}[c]{@{}l@{}}Share of\\Revenue\end{tabular} & nRMSE & \begin{tabular}[c]{@{}l@{}}WnRMSE \end{tabular} \\ 
\midrule
Significant  cluster & >2000                                                        & 0.90                                                    & 1.04  & 0.52                                                                                                \\
Top 100     & 100                                                         & 0.37                                                    & 0.34  & 0.30                                                                                         \\
101-500     & 400                                                         & 0.30                                                    & 0.56  & 0.51                                                                     \\
501-1000    & 500                                                         & 0.12                                                    & 0.75  & 0.69                                                       \\
Above 1001  & >1000                                                       & 0.11                                                    & 1.32  & 0.99                                                          \\ \bottomrule
\end{tabular}
\end{table}

To do further analysis, we ranked the entire dataset based on weight contribution. A significant O\&D cluster is a subset of the dataset that contributes to 90\% of the total revenue across the airline network in the year before the validation period. This significant cluster contains 2000+ O\&D pairs. Aggregated metrics are reported in Table \ref{table2}. Both RMSE and WnRMSE indicate performance degrades towards the tail end of the ranked O\&D. Additionally, the significant cluster has low WnRMSE. This implies that the forecast can capture trends where variability averages out due to volume as compared to a more stochastic tail end. In table \ref{table3}, we use the same dataset slices and evaluate the winning ratios of the machine learning methods as compared to (a) statistical baseline and (b) industry benchmark i.e. year-over-year predictions. Although, the winning percentage for machine learning models is about 50\% as compared to the baseline, we outperformed the industry benchmark overall. Hence, the utility of a mixture of experts is relevant as it provides a fallback option where machine learning models are not performing better than the statistical baselines. 

\begin{table}[H]
\caption{Win ratios of ML models for clustered O\&Ds}
\label{table3}
\begin{tabular}{@{}llllll@{}}
\toprule
O\&D's  &  \begin{tabular}[c]{@{}l@{}}Statistical\\ baseline\end{tabular}  & \begin{tabular}[c]{@{}l@{}}Industry\\ benchmark\end{tabular} \\ 
\midrule
Significant cluster  & 0.49     & 0.94                                            \\
Top 100     & 0.42      & 0.96                                              \\
101-500     & 0.49    & 0.96                                                \\
501-1000    & 0.46   & 0.93                                              \\
Above 1001  & 0.50      & 0.94                                              \\ \bottomrule
\end{tabular}
\end{table}

\section{Future work}
In the future, we intend to extend this work to other target variables that are as crucial as shipped weight for the airline in its decision-making. These include shipped volume (in $m^3$), chargeable weight (normalized metric between gross weight and volumetric weight), yield, and/or revenue.

Furthermore, the proposed mixture of experts framework currently misses out on integrating information from models other than the best performing one on a particular O\&D pair, limiting its predictive power. Therefore, we propose extending the framework to a two-step forecasting process. In the first step, we train simpler models such as statistical benchmarks, decision stumps, and simple DNN networks. The predictions from these models would serve as additional features for advanced deep learning models like NBEATS and TFT, enhancing overall forecast accuracy. Integrating information from these simpler models allows the deep learning model to focus on complex, context-driven scenarios without being burdened with learning the simple trend and seasonality patterns already captured by the statistical models in the first step.

Lastly, we would like to build an attribution system to measure the revenue uplift from decisions made using improved forecasting techniques. Thereby, validating the overall approach for time series forecast in the air cargo industry.

\section{Conclusion}

The field study presented in this paper demonstrates the efficacy of advanced machine learning models in improving mid-term demand forecasting for the air cargo industry. 
Our experiments show that machine learning models perform better than statistical baseline models. Additionally, mixture of experts which combines both statistical and machine learning models, significantly outperforms machine learning only approaches.
The real-world application of these models within a major air cargo carrier underscores their practical utility and potential for broader adoption in other volatile market environments. Future research will focus on extending this framework to additional target variables and enhancing the predictive power by incorporating a two-step forecasting process. This work lays a strong foundation for further advancements in the adoption of machine learning techniques in air cargo demand forecasting, ultimately contributing to more efficient and informed decision-making processes in the industry.




\begin{acks}
We sincerely and gratefully acknowledge our airline partners for
their continuing support.
\end{acks}

\bibliographystyle{ACM-Reference-Format}
\bibliography{main}










\end{document}